\documentclass{article}
\usepackage[preprint]{log_2024}			% for preprint version

\usepackage{booktabs}						% professional-quality tables
\usepackage{multirow}						% tabular cells spanning multiple rows
\usepackage{amsfonts}						% blackboard math symbols
\usepackage{graphicx}						% figures
\usepackage{subcaption}
\usepackage{duckuments}						% sample images
\usepackage{comment}
\usepackage{algpseudocode}
\usepackage{algorithm}
\usepackage{multirow}
\usepackage{colortbl}
\usepackage{amssymb}
\usepackage{times}
\usepackage{epsfig}
\usepackage{xcolor}
\usepackage{boldline}

\usepackage[numbers,compress,sort]{natbib}	% for numerical citations
\title[Multi-Scale High-Resolution Logarithmic Grapher Module for Efficient Vision GNNs]{Multi-Scale High-Resolution Logarithmic Grapher Module for Efficient Vision GNNs}

\author[M. Munir et al.]{%
Mustafa Munir\\
The University of Texas at Austin\\
\email{mmunir@utexas.edu}\And
Alex Zhang\\
The University of Texas at Austin\\
\email{alex.zhang@utexas.edu}\And
Radu Marculescu\\
The University of Texas at Austin\\
\email{radum@utexas.edu}
}

\begin{document}

\maketitle

\begin{abstract}
Vision graph neural networks (ViG) have demonstrated promise in vision tasks as a competitive alternative to conventional convolutional neural nets (CNN) and transformers (ViTs); however, common graph construction methods, such as k-nearest neighbor (KNN), can be expensive on larger images. While methods such as Sparse Vision Graph Attention (SVGA) have shown promise, SVGA’s fixed step scale can lead to over-squashing and missing multiple connections to gain the same information that could be gained from a long-range link. Through this observation, we propose a new graph construction method, Logarithmic Scalable Graph Construction (LSGC) to enhance performance by limiting the number of long-range links. To this end, we propose LogViG, a novel hybrid CNN-GNN model that utilizes LSGC. Furthermore, inspired by the successes of multi-scale and high-resolution architectures, we introduce and apply a high-resolution branch and fuse features between our high-resolution and low-resolution branches for a multi-scale high-resolution Vision GNN network. Extensive experiments show that LogViG beats existing ViG, CNN, and ViT architectures in terms of accuracy, GMACs, and parameters on image classification and semantic segmentation tasks. Our smallest model, Ti-LogViG, achieves an average top-1 accuracy on ImageNet-1K of 79.9\% with a standard deviation of $\pm$ 0.2\%, 1.7\% higher average accuracy than Vision GNN with a 24.3\% reduction in parameters and 35.3\% reduction in GMACs. Our work shows that leveraging long-range links in graph construction for ViGs through our proposed LSGC can exceed the performance of current state-of-the-art ViGs. Code is available at \url{https://github.com/mmunir127/LogViG-Official}.
\end{abstract}

\section{Introduction}
The meteoric rise of deep learning over the past decade has resulted in numerous successes in computer vision. From the development of AlexNet \cite{Alexnet2012} and convolutional neural networks (CNNs) \cite{howard2017mobilenetsefficientconvolutionalneural, Resnet, xu2021regnetselfregulatednetworkimage, tan2021efficientnetv2smallermodelsfaster} to a new generation of vision Transformers (ViTs) \cite{dosovitskiy2021imageworth16x16words, Deit,  bao2022beitbertpretrainingimage},  neural networks have demonstrated their effectiveness in computer vision across the board. In the same vein, CNNs and ViTs have also shown competitive performance toward dense prediction tasks, such as semantic segmentation \cite{ronneberger2015unetconvolutionalnetworksbiomedical, zhang2022segvitsemanticsegmentationplain}. More recently, Graph neural networks (GNNs), specifically Vision GNNs (ViGs) have emerged as competitive alternatives to current CNN and ViG models in computer vision \cite{Vision_GNN, MobileViG, ViHGNN}. Instead of a sliding window over a grid of pixels as in CNNs or a sequence of patches in ViTs, ViGs represent an image as a network of patches linked by content rather than spatial position \cite{Vision_GNN}. Through these patches, a ViG can identify objects within an image by relating each patch with its neighbor. For example, if one patch contains features that are associated with a tire and another patch, connected with the first, contains features that represent handlebars, then a ViG is capable of generalizing the object as a bicycle. Even if two patches are spatially distant, ViGs can exploit long-range links to form the shortest path between spatially distant patches.

Despite promising results, ViGs are computationally complex, especially during network generation \cite{MobileViG}. A common method is to use K nearest neighbors (KNN) and generate a network by selecting a patch and connecting each patch to the K most similar patches \cite{Vision_GNN}. Yet, as mentioned earlier, this method is computationally inefficient, especially for high-resolution images as the time needed to generate a graph can grow exponentially depending on the size of the image. An alternative is a static, structured grapher (SVGA) \cite{MobileViG} which grows linearly horizontally and vertically across an image. While this significantly simplifies the process, this method suffers from over-squashing where too much information is being assimilated into a single vector. This becomes more apparent as the resolution increases, causing the number of connected patches to grow quadratically. To address this \cite{MobileViGv2} proposed a fixed number of connections regardless of input resolution, but this can cause ViGs to lose their global context in high-resolution images.

To address these issues, we propose Logarithmic Scalable Graph Construction (LSGC) as an efficient alternative to KNN and SVGA-style ViGs. LSGC exploits logarithmic growth to create networks that scale with image size, simultaneously avoiding over-squashing and reducing computational complexity. We also deploy LSGC in a hybrid ViG-based CNN-GNN architecture, LogViG, and utilize a high-resolution shortcut \cite{HRViT} to inject high-resolution features at later stages into the model. We summarize our contributions as follows:

\begin{itemize}
    \item We propose a novel approach to graph construction designed for efficient Vision GNNs. Instead of scaling the graph linearly as in SVGA, we scale the graph logarithmically. Logarithmic Scalable Graph Construction (LSGC) improves on SVGA in several ways. First, it generates fewer connections for high-resolution images to help mitigate over-squashing. Second, it helps preserve locality by prioritizing connections closer to a patch without sacrificing the ability to make long-range connections.
    \item We introduce a novel CNN-GNN architecture, LogViG, for image classification and semantic segmentation that uses LSGC and a high-resolution skip connection. We utilize convolutional and grapher layers in all four stages, performing local and global processing at each stage.
    \item We demonstrate that LoGViG outperforms traditional CNN, ViT, and ViG architectures on vision tasks and that LSGC broadly outperforms SVGA in image classification and semantic segmentation.  

\end{itemize}

The paper is arranged as follows. Section 2 covers related work in the ViG and efficient computer vision architecture space. Section 3 describes the design methodology behind LSGC and the LogViG architecture. Section 4 describes the experimental setup and results for ImageNet-1k \cite{imagenet1k} image classification and ADE20K \cite{ADE20K} semantic segmentation. Section 5 covers ablation studies on how different design decisions impact performance on ImageNet-1k. Lastly, Section 6 summarizes our main contributions.

\section{Related Work}
Current network architectures for computer vision commonly utilize convolution neural networks (CNNs) \cite{howard2017mobilenetsefficientconvolutionalneural, Resnet, xu2021regnetselfregulatednetworkimage, tan2021efficientnetv2smallermodelsfaster} and vision Transformers (ViTs) \cite{dosovitskiy2021imageworth16x16words, Deit, bao2022beitbertpretrainingimage}. On dense prediction tasks, ViTs tend to outperform CNNs \cite{jeeveswaran2022comprehensivestudyvisiontransformers}; however, are computationally complex compared to CNNs, resulting in higher compute times and latency \cite{thisanke2023semanticsegmentationusingvision, 10.1145/3508396.3512869}. In addition, ViT performance degrades on high-resolution images \cite{jeeveswaran2022comprehensivestudyvisiontransformers}. Conversely, CNNs lack a global receptive field and cannot capture global features when compared with ViTs \cite{jeeveswaran2022comprehensivestudyvisiontransformers}. In both cases, CNNs and ViTs are limited in their ability to represent image data, restricted to a grid of pixels or a sequence of patches respectively \cite{Vision_GNN}. While CNNs tend to be more efficient than ViTs \cite{10.1145/3508396.3512869}, some headway has been made to make ViTs competitive \cite{liu2023efficientvitmemoryefficientvision}. Still, pure CNN and ViT models suffer from the drawbacks discussed, opening the door for alternative architectures. 

Vision GNNS are a proposed alternative to CNN and ViT-based architectures. Traditional ViG networks represent images by computing the k nearest-neighbors (KNN) for every patch in an image \cite{Vision_GNN}, attending to similar patches across an entire image. The result is a network of short and long-range connections across the image. By representing an image as a network of patches, it bypasses the inflexibility issues of CNNs and ViTs. Additionally, the mixture of short and long-range connections between patches allows ViGs to capture both local and global features, a trait lacking in CNNs.

Historically, graph neural networks (GNNs) have been used for biological, language, and social data \cite{ma2018affinitynetsemisupervisedfewshotlearning, marcheggiani2017encodingsentencesgraphconvolutional, Nguyen_Grishman_2018, Qiu_2018}. GNNs have also been used for vision tasks such as object localization, detection, and classification \cite{monfardini2006graph,qi2017pointnetdeeplearningpoint, 9892682}. After the introduction of Vision GNN \cite{Vision_GNN} (ViGs), GNN-based networks have grown dramatically \cite{MobileViG, 10230496,shou2024graphinformationbottleneckremote}. Yet, despite significant progress, graph construction remains computationally expensive \cite{MobileViG}. While static grapher methods such as Sparse Vision Graph Attention (SVGA) help mitigate the computational complexity issue of ViGs, they fail to scale with high-resolution images and introduce over-squashing. To tackle this issue, we introduce Logarthimic Scalable Graph Construction (LSGC), an efficient graph construction method capable of scaling with high-resolution images, and LogViG, a novel CNN-GNN hybrid network utilizing LSGC that outperforms competing state-of-the-art (SOTA) architectures.

\begin{figure}
    \centering
    \begin{subfigure}{.30\linewidth}
        \centering
        \includegraphics[width=0.8\linewidth]{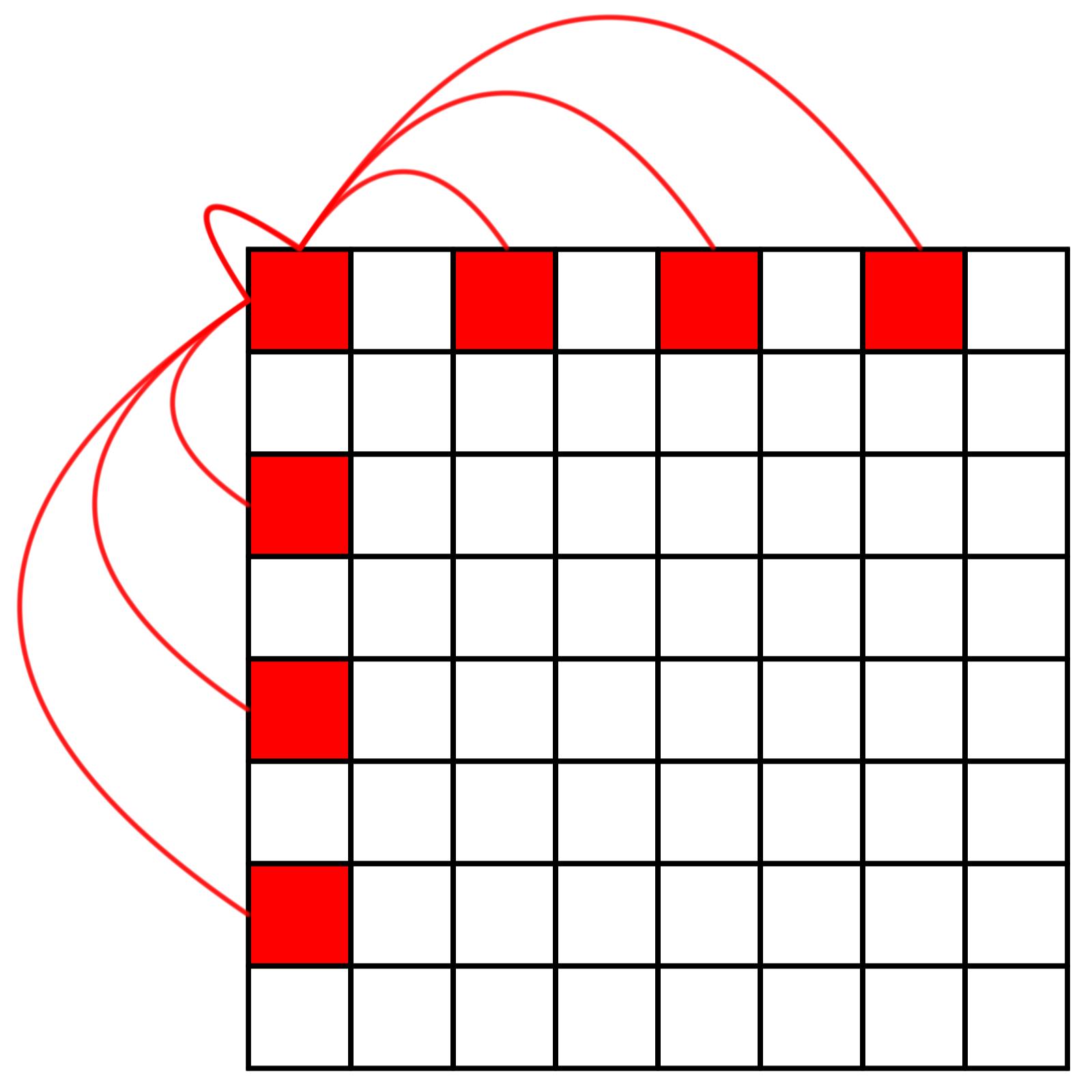}
        \caption{SVGA Grapher}
    \end{subfigure}
    \begin{subfigure}{.30\linewidth}
        \centering
        \includegraphics[width=0.8\linewidth]{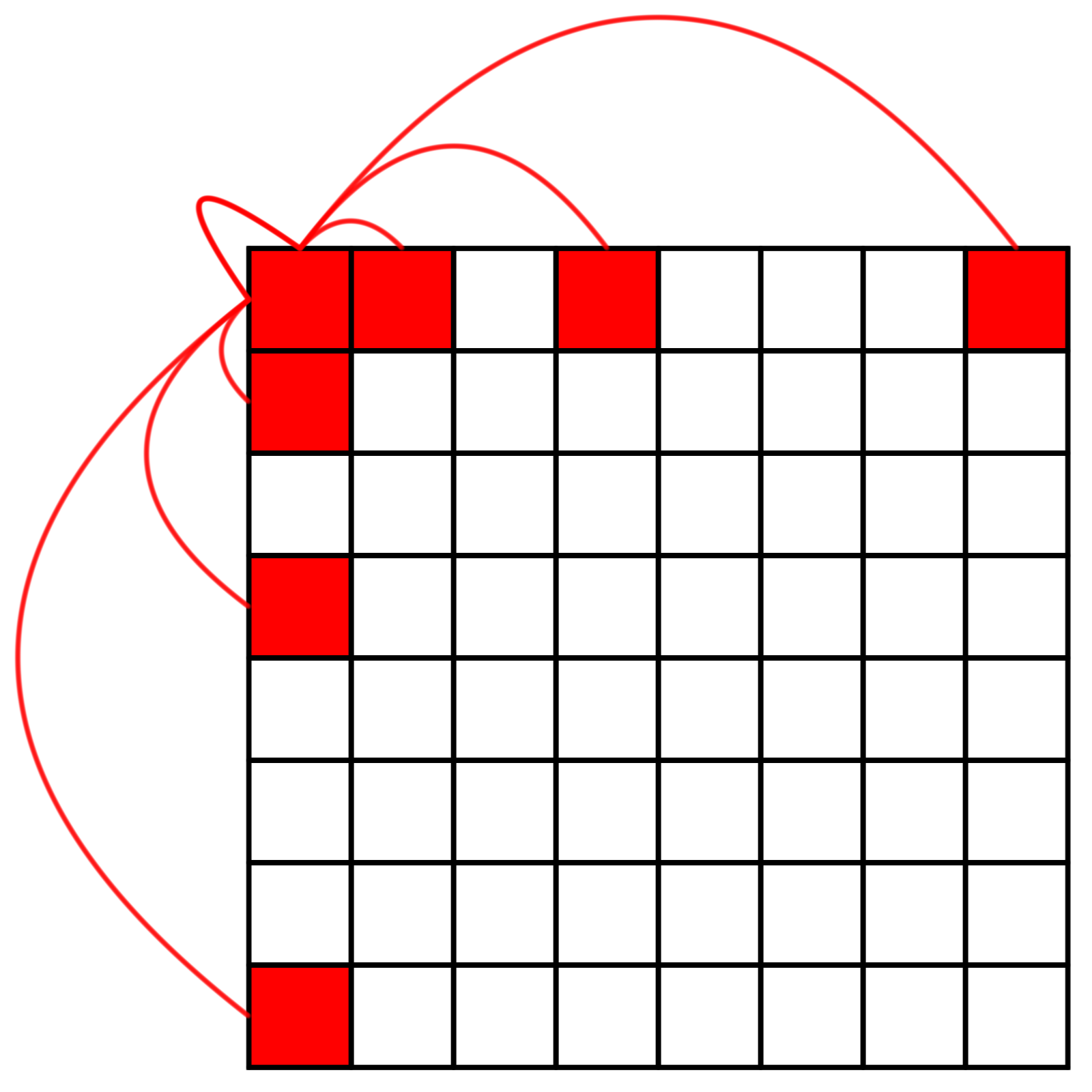}
        \caption{LSGC Forward Pass}
    \end{subfigure}
     \begin{subfigure}{.30\linewidth}
        \centering
        \includegraphics[width=0.8\linewidth]{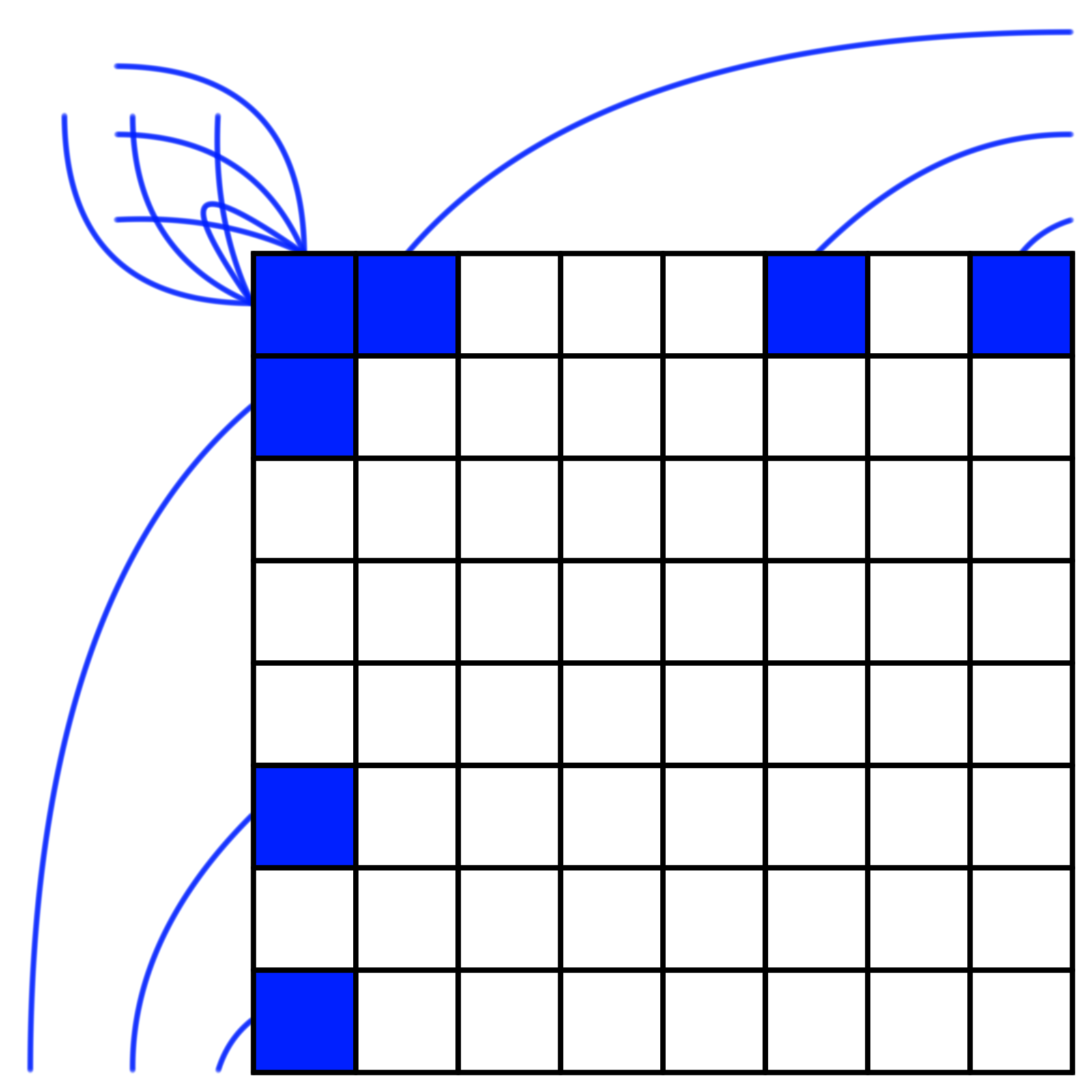}
        \caption{LSGC Backward Pass}
    \end{subfigure}
    \caption{a) SVGA graph attention for the top left pixel of an $8\times8$ image b) LSGC forward graph attention pass for the top left pixel of an $8\times8$ image and an expansion rate $K = 2$. As depicted, LSGC grows logarithmically rather than linearly. c) LSGC backward graph attention pass for the top left pixel of an $8\times8$ image and expansion rate $K = 2$. As shown, the LSGC wraps around the image to create connections}
    \label{fig:graphers}
\end{figure}

\section{Methodology}
\subsection{Logarithmic Scalable Graph Construction}

We propose Logarithmic Scalable Graph Construction (LSGC) as an alternative efficient graph construction method to KNN graph attention from Vision GNN (ViG) \cite{Vision_GNN}. We expand upon SVGA \cite{MobileViG} by scaling the graph logarithmically based on the bit-length of an image rather than statically. In this way, the grapher is efficient on large images by constructing fewer links than SVGA while still creating global links over the entire image. In addition, these fewer, but global, connections avoid over-squashing \cite{alon2020bottleneck} and redundancy by limiting the number of links per image patch.

For ViG architectures, the k-Nearest-Neighbors (KNN) computation is required for every input image, since one cannot know the nearest neighbors of every pixel on an image. This results in a graph with connections scattered over the image. Due to the nature of KNN, ViG contains two reshaping operations \cite{Vision_GNN}. The first reshapes the input image from a 4D tensor to a 3D tensor for graph convolution and the second reshapes the input from 3D back to 4D for convolutional layers. SVGA \cite{MobileViG} eliminates these reshaping operations utilizing a static graph where each patch is connected to every $K^{th}$ patch in its row and column as in Figure~\ref{fig:graphers}a. A follow-up work to MobileViG \cite{MobileViG} proposes MGC \cite{MobileViGv2} which utilizes a fixed number of possible connections per token regardless of input size.

LSGC replaces the static graph structure of SVGA or the fixed number of connections of MGC and improves upon them with a logarithmic structure that scales based on the input image size. To do this, we first obtain the bit-depth from $H$ and $W$, the height and width of the input image in pixels. We define bit-depth as the number of bits required to represent the dimensions in binary. For example, an $8\times8$ input results in a bit-depth of four for both width and height. We reason that this will ensure a global connection can be established for larger images because any pixel will be able to at least reach across an image to establish a long-distance connection. After calculating the bit-depth, we store these values as $h$ and $w$ as denoted in Algorithm~\ref{alg:LSGC}.

Next, we implement graph construction using a series of $expand$ operations which we denote as $expand_{forward}$ and $expand_{backward}$ as described in Algorithm~\ref{alg:LSGC}. Using the bit-depth $h$ and $w$ as our scaling limit, and setting the expansion rate $K$, we first $expand_{forward}$ in the downwards and right directions (denoted as $expand_{forwardH}$ and $expand_{forwardW}$) from a pixel and create connections every $2^n - 1\  \forall\ 1 \leq n \leq h,w$ pixels as shown for an $8\times8$ image in Figure~\ref{fig:graphers}b. Similarly, we then $expand_{backward}$ in the upwards and left directions (denoted as $expand_{backwardH}$ and $expand_{backwardW}$) opposite of $expand_{forward}$. If there is no room to expand, as our example in Figure~\ref{fig:graphers}c demonstrates, we simply wrap the image around such that we expand to the other side of the image. Then, after every directional expansion, we perform max-relative graph convolution (MRConv). To do this, we compute the element-wise $max$ operation over the difference between the original image $X$ and the expanded version computed by the $expand$ operation and store the result in $X_j$ as shown in Algorithm~\ref{alg:LSGC}. Finally, we concatenate $X_j$ with the original image and apply a $Conv2d$ over the entire matrix. In this way, we achieve reduced computational complexity for graph construction compared to SVGA and KNN whilst still establishing global connections throughout the image. The effect of expansion is demonstrated in Table ~\ref{tab:AvgPath}. While a normal square lattice has a significantly larger shortest path length across the image, LSGC allows for links to be established across an entire image with similar effectiveness to SVGA. For example, at the highest resolution of 56 $\times$ 56, a regular square lattice has an average shortest path length of 37.333. After applying LSGC or SVGA, the shortest path length is reduced to 4.359 and 2.895 respectively as shown in Table ~\ref{tab:AvgPath}. The reduction of shortest path length indicates that longer-range links are being established, permitting global data over the entire image to be seen and processed. We note that SVGA has a shorter average path length than LSGC due to making significantly more connections, which leads to the over-squashing \cite{alon2020bottleneck} effect SVGA suffers from.

\begin{table}
	\centering
	\caption{Shortest Average Path on Various Resolutions and Network Structures.}
	\label{tab:AvgPath}
	\begin{tabular}{cccc}
		\toprule
		Resolution & Lattice & LSGC & SVGA  \\
		\midrule
        56 X 56 & 37.333 & 4.359 & 2.895 \\
        28 X 28 & 18.667 & 3.719 & 2.794 \\
        14 X 14 & 9.333 & 3.303 & 2.605 \\
        7 X 7 & 4.667 & 2.334 & 1.750 \\
		\bottomrule
	\end{tabular}
\end{table}

\begin{algorithm}
    \caption{LSGC}
    \begin{algorithmic}
    \Require $H, W$, the image resolution; $X$, the input image; $K$ the expansion rate
    
    \State $h \gets bit depth(H)$ \Comment{Calculate bit depth}
    \State $w \gets bit depth(W)$ 
    \State $X_j \gets 0$
    \For {$i = 1,2,\dots$  $h$}
    \State $X_c \gets X - expand_{forwardH} (X, K^i - 1)$  \Comment{Get downwards relative features}
    \State $X_j \gets max(X_c, X_j)$                      \Comment{Keep max relative features}
    \EndFor
    
    \For {$i = 1,2,\dots$  $w$}
    \State $X_c \gets X - expand_{forwardW} (X, K^i - 1)$
    \State $X_j \gets max(X_c, X_j)$
    \EndFor
    
    \For {$i = 1,2,\dots$  $h$}
    \State $X_c \gets X - expand_{backwardH} (X, K^i - 1)$
    \State $X_j \gets max(X_c, X_j)$
    \EndFor
    
    \For {$i = 1,2,\dots$  $w$}
    \State $X_c \gets X - expand_{backwardW} (X, K^i - 1)$
    \State $X_j \gets max(X_c, X_j)$
    \EndFor
    
    \State \textbf{return} $Conv2d(Concat(X, X_j)$
    \end{algorithmic}
    \label{alg:LSGC}
\end{algorithm}

\begin{figure}
    \centering
    \includegraphics[width=.85\linewidth]{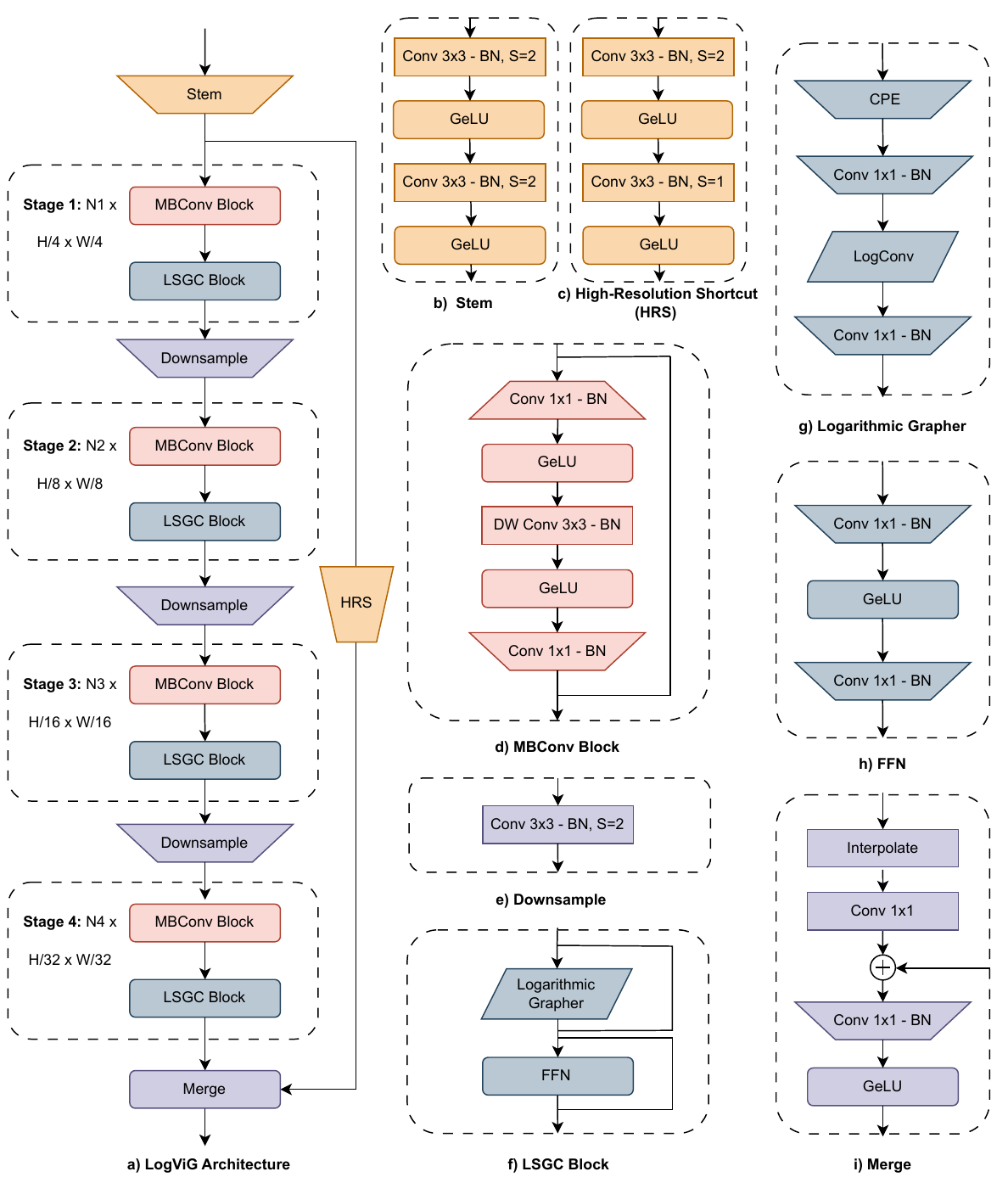}
    \caption{LogViG Architecture. (a) Network architecture showing the stages and blocks. (b) Stem. (c) High-Resolution Shortcut (HRS). (d) MBConv Block. (e)Downsample. (f) LSGC Block. (g) Logarithmic Grapher. (h) FFN. (i) Merge.}
    \label{fig:arch}
\end{figure}

\subsection{High-Resolution Shortcut}

Networks typically downsample the feature maps to operate on lower-resolution representations due to the associated computational complexity (increased GMACs) with operating on high-resolution images \cite{HRViT, HRNet}. The issue with such networks is that they lack cross-resolution interaction, which can lead to worse performance.
Inspired by prior work such as HRNet \cite{HRNet}, Lite-HRNet \cite{Lite_HRNet}, and HRViT \cite{HRViT} we propose to bring multi-scale high-resolution interactions to Vision GNNs.

To enable multi-scale feature interaction we use a High-Resolution Shortcut (HRS), which allows us to merge our higher resolution features with our lower resolution features. Our HRS consists of two $3\times3$ convolutions, one with a $stride=2$ and the other with a $stride=1$, each followed by batch normalization (BN) and the GeLU activation function as shown in Figure \ref{fig:arch}c. We then merge the features of our high-resolution branch with our low-resolution branch. To merge the features, first we upsample our low-resolution features through bilinear interpolation. Next, we perform a pointwise convolution to match the channel dimensions. Lastly, we sum our features, pass them into another pointwise convolution, followed by BN and GeLU.

\subsection{Going Deeper Versus Going Wider}

Deeper networks have been shown to better learn hierarchical features, which can improve accuracy on complex and large datasets \cite{nguyen2020wide}. To optimize our proposed LogViG architecture we studied whether our network performance is improved with a deeper and narrower network or a shallower and wider network. Namely, we compared our Ti-LogViG architecture shown in Table \ref{tab:table_of_arch} with the Wide Ti-LogViG architecture. For our wider variation we cut the depth by 33\% and increased the channel width by 50\% for stages 1 to 3. For stage 4 we increased the channel width to 384 from 224 to get a similar number of parameters to our Ti-LogViG so we could fairly compare the deeper and wider network. The results in Table \ref{tab:ablation_1} show our comparison of both networks and we find the deeper LogViG-Ti achieves a 0.4\% accuracy increase with only a 0.1 M increase in parameters.

\subsection{Network Architecture}

The overall LogViG architecture is shown in Figure \ref{fig:arch}a. First, we pass the image through the convolutional stem. The stem takes the input images and downsamples it 4$\times$ using convolutions of $stride = 2$. The output of the stem is fed into two branches: The low-resolution branch and the high-resolution branch. In the low-resolution branch, the stem output is passed through four stages. Each stage is made up of two blocks: the MBConv Block and LSGC Block as described in Figure \ref{fig:arch}a. Between each stage are additional convolution-based downsampling steps. The high-resolution branch is made up of a single High-Resolution Shortcut block as shown in \ref{fig:arch}a. The output of the low-resolution branch is interpolated and channel adjusted before being added to the high-resolution branch as demonstrated in Figure \ref{fig:arch}i. Lastly, the merge block output is put through an average pooling step followed by a feed-forward network, producing the predicted class of the input image.

\begin{table}[H]
\small
\caption{\textbf{Architecture details of LogViG} showing configuration of the stem, stages, and classification head. $C$ represents the channel dimensions.}
\centering
\setlength{\tabcolsep}{6pt}
\begin{tabular}{|c|c|c|c|c|c|c|}
\hline
Stage                             & Ti-LogViG & S-LogViG & B-LogViG & Wide Ti-LogViG \\ \hline \rule{0pt}{4ex}
Stem                                    & Conv $\times$2             & Conv $\times$2            & Conv $\times$2        &   Conv $\times$2  \\[8pt] \hline \rule{0pt}{4ex}
Stage 1           & $ \begin{array}{ccc} MBConv \times3 \\ LSGC \times3 \\ C = 32 \end{array} $             & $ \begin{array}{ccc} MBConv \times5 \\ LSGC \times5 \\ C = 32 \end{array} $          
& $ \begin{array}{ccc} MBConv \times5 \\ LSGC \times5 \\ C = 48 \end{array} $  & $ \begin{array}{ccc} MBConv \times1 \\ LSGC \times1 \\ C = 48 \end{array} $
 \\[8pt] \hline \rule{0pt}{4ex}           
Stage 2           & $ \begin{array}{ccc} MBConv \times3 \\ LSGC \times3 \\ C = 64 \end{array} $             & $ \begin{array}{ccc} MBConv \times5 \\ LSGC \times5 \\ C = 64  & \end{array} $          
& $ \begin{array}{ccc} MBConv \times5 \\ LSGC \times5 \\ C = 96 \end{array} $ & $ \begin{array}{ccc} MBConv \times1 \\ LSGC \times1 \\ C = 96 \end{array} $ 
\\[8pt] \hline \rule{0pt}{4ex}
Stage 3              & $ \begin{array}{ccc} MBConv \times9 \\ LSGC \times3 \\ C = 128 \end{array} $             & $ \begin{array}{ccc} MBConv \times15 \\ LSGC \times5 \\ C = 128 & \end{array} $          
& $ \begin{array}{ccc} MBConv \times15 \\ LSGC \times5 \\ C = 192 \end{array} $ & $ \begin{array}{ccc} MBConv \times3 \\ LSGC \times1 \\ C = 192 \end{array} $
\\[8pt] \hline \rule{0pt}{4ex}
Stage 4                   & $ \begin{array}{ccc} MBConv \times3 \\ LSGC \times3 \\ C = 224 \end{array} $             & $ \begin{array}{ccc} MBConv \times5 \\ LSGC \times5 \\ C = 256 \end{array} $          
& $ \begin{array}{ccc} MBConv \times5 \\ LSGC \times5 \\ C = 384 \end{array} $ & $ \begin{array}{ccc} MBConv \times1 \\ LSGC \times1 \\ C = 384 \end{array} $       \\[8pt] \hline 
Head                                         & Pooling \& MLP            & Pooling \& MLP            & Pooling \& MLP     &   Pooling \& MLP     \\ \hline
\end{tabular}
\label{tab:table_of_arch}
\end{table}

The network architecture details for Ti-LogViG, S-LogViG, and B-LogViG are provided in Table \ref{tab:table_of_arch}. We also provide the details of Wide Ti-LogViG, which we used to determine the improvement in performance we would gain by going deeper versus wider in our network. We report the configuration of the stem, stages, and classification head. For each stage we also report the channel dimensions and the number of MBConv and LSGC blocks used.

\section{Experimental Results}

We compare LogViG to ViG\cite{Vision_GNN} and MobileViG \cite{MobileViG} to show its superior performance in terms of image classification accuracy on ImageNet-1k\cite{imagenet1k} in Table \ref{tab:Classification_Results} for all model sizes. We also demonstrate superior performance in terms of semantic segmentation on ADE20K \cite{ADE20K} for all model sizes compared in Table \ref{tab:Segmentation_Results}. For each SOTA model we compare to \textbf{LogViG obtains superior performance for similar or less parameters and GMACs}.

\definecolor{LightGray}{gray}{0.7} % Define a light gray color; adjust 0.9 to your liking

\begin{table*}[ht]
\fontsize{10.5}{10.5}\selectfont
\def\arraystretch{1.2}
\caption{\textbf{Classification results on ImageNet-1k} for LogViG and other state-of-the-art models. Bold entries show best results for competing models for each column. Gray highlighted entries indicate results obtained for LogViG proposed in this paper. The Top-1 accuracy results for LogViG models are averaged over four experiments and show the mean $\pm$ standard deviation.}
\centering
\begin{tabular}[t]{|c|c|c|c|c|c|c|c|c|}
\hline
{\textbf{Model}} & {\textbf{Type}} & {\textbf{Params (M)}} & {\textbf{GMACs}} & {\textbf{Epochs}} & {\textbf{Acc (\%)}}        \\ \hline

ResNet18 \cite{Resnet} & CNN & 11.7 & 1.82 & 300 & 69.7 \\ \hline
ResNet50 \cite{Resnet}                       & CNN          & 25.6      & 4.1       & 300 & 80.4      \\ \hline
ConvNext-T \cite{liu2022convnet}                     & CNN          & 28.6      & 7.4        & 300 & 82.7 

\\ \hlineB{5}
EfficientFormer-L1 \cite{EfficientFormer}             & CNN-ViT        & 12.3      & 1.3      & 300 & 79.2     \\ \hline
EfficientFormer-L3 \cite{EfficientFormer}              & CNN-ViT        & 31.3      & 3.9    & 300 & 82.4      \\ \hline
EfficientFormer-L7 \cite{EfficientFormer}              & CNN-ViT        & 82.1      & 10.2   & 300 & 83.3 
    \\ \hline
LeViT-192 \cite{graham2021levit}            & CNN-ViT        & 10.9      & \textbf{0.7}       & 1000 & 80.0     \\ \hline
LeViT-384 \cite{graham2021levit}                      & CNN-ViT        & 39.1      & 2.4        & 1000 & 82.6     \\ \hlineB{5}
HRViT-b1 \cite{HRViT}       & ViT    & 19.7     & 2.7      & 300    & 80.5 \\ \hline
HRViT-b2 \cite{HRViT}       & ViT    & 32.5     & 5.1      & 300    & 82.3 \\ \hline
HRViT-b3 \cite{HRViT}             & ViT        & 37.9      & 5.7      & 300 & 82.8          \\ \hline
PVT-Small \cite{wang2021pyramid}             & ViT         & 24.5          & 3.8       & 300 & 79.8 \\ \hline
PVT-Large \cite{wang2021pyramid}             & ViT         & 61.4          & 9.8       & 300 & 81.7 \\ \hline
DeiT-S \cite{Deit}                         & ViT     & 22.5      & 4.5       & 300 & 81.2      \\ \hline
Swin-T \cite{liu2021swin}                      & ViT     & 29.0      & 4.5        & 300 & 81.4      \\ \hline
PoolFormer-S12 \cite{MetaFormer}            & Pool          & 12.0      & 2.0        & 300 & 77.2     \\ \hline
PoolFormer-S24 \cite{MetaFormer}                 & Pool          & 21.0      & 3.6        & 300 & 80.3      \\ \hline 
PoolFormer-S36 \cite{MetaFormer}                 & Pool          & 31.0      & 5.2         & 300 & 81.4       \\ \hlineB{5}
PViHGNN-Ti  \cite{ViHGNN}       & GNN & 12.3     & 2.3 & 300 & 78.9       \\ \hline
PViHGNN-S    \cite{ViHGNN}     & GNN & 28.5     & 6.3 & 300 & 82.5      \\ \hline
PViHGNN-B    \cite{ViHGNN}     & GNN & 94.4 & 18.1     & 300 & \textbf{83.9}    \\ \hline
PViG-Ti \cite{Vision_GNN}          & GNN & 10.7     & 1.7 & 300 & 78.2       \\ \hline
PViG-S \cite{Vision_GNN}          & GNN & 27.3     & 4.6 & 300 & 82.1      \\ \hline
PViG-B \cite{Vision_GNN}          & GNN & 92.6 & 16.8     & 300 & 83.7     \\ \hlineB{5}
{MobileViG-S} \cite{MobileViG}    & {CNN-GNN}        & {\textbf{7.2}}       & {1.0}     & {300} & {78.2}       \\ \hline
{MobileViG-M} \cite{MobileViG}     & {CNN-GNN}        & {14.0}         & {1.5}   & {300} & {80.6}       \\ \hline
{MobileViG-B}  \cite{MobileViG}    & {CNN-GNN}        & {26.7}         & {2.8}     & {300} & {82.6}       \\ \hlineB{5}
\rowcolor {LightGray}
Ti-LogViG (Ours)     & CNN-GNN        & 8.1       & 1.1     & 300 & 79.9 $\pm$ 0.2        \\ \hline
\rowcolor {LightGray}
S-LogViG (Ours)     & CNN-GNN        & 13.9         & 1.9   & 300 & 81.5 $\pm$ 0.1       \\ \hline
\rowcolor {LightGray}
B-LogViG (Ours)     & CNN-GNN        & 30.5         & 4.6     & 300 & 83.6 $\pm$ 0.1       \\ \hlineB{5}
\end{tabular}
\label{tab:Classification_Results}
\end{table*}

\subsection{Image Classification}
We implement the model using PyTorch 1.12 \cite{paszke2019pytorch} and Timm library \cite{timm}. The models are trained from scratch for 300 epochs on ImageNet1K \cite{imagenet1k} with AdamW optimizer \cite{AdamW}. We set the learning rate to 2$e^{-3}$ with a cosine annealing schedule. We use a standard image resolution of 224 $\times$ 224, for both training and testing. Following prior work \cite{Deit, MobileViG, EfficientFormer, MobileViGv2, GreedyViG, EfficientFormerv2}, we perform knowledge distillation using RegNetY-16GF \cite{RegNetY} with 82.9\%  top-1 accuracy. For data augmentation we use RandAugment \cite{RandAugment}, Mixup\cite{Mixup}, Cutmix\cite{CutMix} random erasing \cite{RandomErase}, and repeated augment \cite{RepeatedAugment}.

As seen in Table~\ref{tab:Classification_Results}, for a similar number of parameters and GMACs, LogViG outperforms high-resolution architectures such as HRViT \cite{HRViT}, as well as GNN architectures such as Pyramid ViG (PViG) \cite{Vision_GNN}, Pyramid ViHGNN (PViHGNN) \cite{ViHGNN}, and MobileViG \cite{MobileViG} by a significant margin. For example, our S-LogViG, achieves 81.5\%  top-1 accuracy on ImageNet-1K with 13.9 million (M) parameters and 1.9 GMACs compared to HRViT-b2 \cite{HRViT} with 80.5\% top-1 accuracy at 19.7 M parameters and 2.7 GMACs. Our largest model B-LogViG obtains 83.6\% top-1 accuracy with only 30.5 M parameters and 4.6 GMACs. Meanwhile, HRViT-b3 \cite{HRViT} obtains only 82.8\%, a 0.8\% decrease in accuracy, at a higher cost of 37.9 M parameters and 5.7 GMACs.

Compared to similar GNN architectures, our smallest model, Ti-LogViG outperforms PViHGNN-Ti \cite{ViHGNN} and PViG-Ti\cite{Vision_GNN} in accuracy with less parameters and GMACs. Ti-LogViG obtains 79.9\% top-1 accuracy at 8.1 M parameters and 1.1 GMACs. Meanwhile, PViHGNN-Ti \cite{ViHGNN} only achieves 78.9\% top-1 accuracy at 12.3 M parameters and 2.3 GMACs, a 1.0\% difference in accuracy for 4.3 M more parameters and 1.1 more GMACs. Additionally, PViG-Ti \cite{Vision_GNN} reaches 78.2\% top-1 accuracy at 10.7 M parameters and 1.7 GMACs, a decrease of 1.7\% for higher parameters and GMACs.

When juxtaposed to other architectures in Table ~\ref{tab:Classification_Results}, LogViG beats SOTA models in accuracy for a similar number of parameters and GMACs. Ti-LogViG beats PoolFormer-S12 \cite{MetaFormer} with 2.7\% higher top-1 accuracy and 3.9 M fewer parameters and 0.9 fewer GMACs. S-LogViG beats DeiT \cite{Deit} with 0.3\% higher top-1 accuracy while having 8.6 M fewer parameters and 2.6 fewer GMACs. Additionally, B-LogViG beats the EfficientFormer \cite{EfficientFormer} family of models with significantly fewer parameters.

\subsection{Semantic Segmentation}

We further compare the performance of LogViG on semantic segmentation using the scene parsing dataset, ADE20k \cite{ADE20K}. The dataset contains 20K training images and 2K validation images with 150 semantic categories. We build LogViG with semantic FPN \cite{kirillov2019panoptic} as the segmentation decoder, following the methodologies of \cite{MetaFormer, EfficientFormer, EfficientFormerv2, FastViT, GreedyViG}. The backbone is initialized with pretrained weights on ImageNet-1K and the model is trained for 40K iterations on 8 NVIDIA RTX 6000 Ada generation GPUs. Following the process of prior works in segmentation, we use the AdamW optimizer, set the learning rate as 2 $\times$ 10$^{-4}$ with a poly decay by the power of 0.9, and set the training resolution to 512 $\times$ 512 \cite{EfficientFormer, EfficientFormerv2, MobileViG, GreedyViG}. 

As shown in Table \ref{tab:Segmentation_Results}, S-LogViG outperforms PoolFormer-S12 \cite{MetaFormer}, FastViT-SA12 \cite{FastViT}, EfficientFormer-L1 \cite{EfficientFormer}, and MobileViG-M by 6.9, 6.1, 5.2, and 2.3 $mIoU$, respectively. Additionally, B-LogViG outperforms FastViT-SA36 \cite{FastViT} by 3.9 $mIoU$ with only 0.1 M more parameters. B-LogViG also outperforms EfficientFormer-L3 \cite{EfficientFormer} by 3.3 $mIoU$ with 0.8 M fewer parameters.

\begin{table}[ht]
\small % or \footnotesize, \scriptsize, etc.
\def\arraystretch{1.2}
\caption{\textbf{Semantic segmentation results} of LogViG and other backbones on ADE20K. Bold entries indicate results obtained using LogViG and LSGC proposed in this paper. The * on MobileViG-M indicates the authors do not report semantic segmentation results so we trained the model and report the mIoU we obtained. For all other models that are not our LogViG we report the mIoU's reported in their original papers. The mIoU results for LogViG models are averaged over four experiments and show the mean $\pm$ standard deviation.}
\centering
\setlength{\tabcolsep}{9pt}
\begin{tabular}[t]{|c|c|c|c|c|}
\hline
{\textbf{Backbone}} & {\textbf{Parameters (M)}} & \textbf{$mIoU$} \\ \hline
                    
ResNet18 \cite{Resnet}        & 11.7  & 32.9    \\ \hline
PVT-Tiny \cite{wang2021pyramid}        & 13.2  & 35.7    \\ \hline
EfficientFormer-L1 \cite{EfficientFormer}   & 12.3 &  38.9       \\ \hline
PoolFormer-S12 \cite{MetaFormer}   & 12.0   & 37.2        \\ \hline
FastViT-SA12 \cite{FastViT}  & 10.9  & 38.0       \\ \hline
{MobileViG-M*} \cite{MobileViG}    & {14.0} & 41.8   \\ \hline
\rowcolor {LightGray}
\textbf{S-LogViG (Ours)}    & \textbf{13.9} & \textbf{44.1 $\pm$ 0.6}    \\ \hlineB{5}

ResNet50 \cite{Resnet}            & 25.6  & 36.7    \\ \hline
PVT-Small \cite{wang2021pyramid}        & 24.5  & 39.8    \\ \hline
PVT-Large \cite{wang2021pyramid}        & 61.4  & 42.1   \\ \hline
EfficientFormer-L3 \cite{EfficientFormer}    & 31.3 & 43.5     \\ \hline
EfficientFormer-L7 \cite{EfficientFormer}    & 82.1 &  45.1       \\ \hline
PoolFormer-S24 \cite{MetaFormer}          & 21.0 & 40.3    \\ \hline
PoolFormer-S36 \cite{MetaFormer}          & 31.0 & 42.0    \\ \hline
PoolFormer-M48 \cite{MetaFormer}          & 73.0 & 42.7    \\ \hline
FastViT-SA36 \cite{FastViT}                   & 30.4 & 42.9 \\ \hline
\rowcolor {LightGray}
\textbf{B-LogViG (Ours)}    & \textbf{30.5}   & \textbf{46.8 $\pm$ 0.4}    \\ \hlineB{5}
\end{tabular}
\label{tab:Segmentation_Results}
\end{table}

\subsection{Ablation Studies}

We perform ablation studies to show the benefits of LSGC, our deeper network configuration, and of our high-resolution shortcut (HRS). A summary of these results can be found in Tables \ref{tab:ablation_1} and \ref{tab:ablation_construction}.

\begin{table*}[h]
\small
\def\arraystretch{1.0}
\caption{An ablation study of the effects of higher resolution graphers in our network, and using deeper and narrower networks. A checkmark indicates this component was used in the experiment. A (-) indicates this component was not used. 1-S indicates that graph convolutions were only used in Stage 4 of the model, 3-S indicates that graph convolutions were used in stages 2, 3, and 4, and 4-S indicates that graph convolutions were used in all stages of the network.}
\centering
\begin{tabular}[t]{c|c|c|c|c|c|c|c}
\hline
\multirow{2}{*}{\textbf{Base Model}} & \multirow{2}{*}{\textbf{Params (M)}} & \multirow{2}{*}{\textbf{LSGC}} & \multirow{2}{*}{\textbf{1-S}} & \multirow{2}{*}{\textbf{3-S}} & \multirow{2}{*}{\textbf{4-S}} & \multirow{2}{*}{\textbf{Deep Network}} & \multirow{2}{*}{\textbf{Top-1 (\%)}} \\
                                              &           &     &  &    &  &  \\ \hline

MobileViG-S                     & 7.2  & - & \checkmark & - & - & - & 78.2 \\
Ti-LogViG                     & 7.0  & \checkmark & \checkmark & - & - & \checkmark & 78.8 \\
Ti-LogViG                     & 8.0  & \checkmark & - & \checkmark & - & \checkmark & 79.8 \\
Ti-LogViG                     & 8.0  & \checkmark & - & - & \checkmark & - & 79.6\\
\rowcolor{LightGray}
Ti-LogViG                     & 8.1  & \checkmark & - & - & \checkmark & \checkmark & 79.9  \\ \hline
\end{tabular}
\label{tab:ablation_1}
\end{table*}

Starting with MobileViG-S as a base model for comparison, we first try using our LSGC-style graph construction and our deeper and narrower network architecture. We find our LogViT-Ti in this case outperforms MobileViG-S by 0.6\% with 0.2 M fewer parameters. Next we use graph convolutions in stages 2, 3, and 4 of the model, the resulting model achieves a top-1 accuracy on ImageNet-1K of 79.8\%, 1.0\% higher than with only using graph convolutions in stage 4. Next when we use graph convolutions in all 4 stages of the network we achieve a top-1 accuracy on ImageNet-1K of 79.9\% with 8.1 M parameters. When we modify the network to be wider and less deep as shown in Table \ref{tab:table_of_arch} indicated by Wide Ti-LogViG, we see a decrease in accuracy of 0.3\% with only 0.1 M fewer parameters.

\begin{table*}[h]
\small
\def\arraystretch{1.0}
\caption{An ablation study of the effects of our LSGC graph construction versus SVGA graph construction and of our high-resolution branch using the high-resolution shortcut. A checkmark indicates this component was used in the experiment. A (-) indicates this component was not used.}
\centering
\begin{tabular}[t]{c|c|c|c|c|c}
\hline
\multirow{2}{*}{\textbf{Base Model}} & \multirow{2}{*}{\textbf{Params (M)}} & \multirow{2}{*}{\textbf{SVGA}}  & \multirow{2}{*}{\textbf{LSGC}} & \multirow{2}{*}{\textbf{High-Resolution}} & \multirow{2}{*}{\textbf{Top-1 (\%)}} \\
                                              &            &  &  & \\ \hline

Ti-LogViG                     & 8.0  & \checkmark & - & - &  79.4 \\
Ti-LogViG                    & 8.0  & - & \checkmark & -  & 79.8 \\
\rowcolor{LightGray}
Ti-LogViG                   & 8.1 & - & \checkmark    & \checkmark & \textbf{79.9}   \\ \hline
\end{tabular}
\label{tab:ablation_construction}
\end{table*}

Starting with Ti-LogViG with SVGA and without our high-resolution shortcut (HRS) we achieve a Top-1 accuracy on ImageNet-1k of 79.4\%. When we add our LSGC we gain 0.4\% in accuracy with no additional parameters. When we add HRS we gain another 0.1\% in accuracy with only 0.1 M additional parameters as shown in Table \ref{tab:ablation_construction}.

\vspace{-2mm}

\section{Conclusion}

In this work, we have introduced Logarithmic Scalable Graph Construction (LSGC), a novel method for constructing graphs in Vision GNNs that efficiently balances the inclusion of long-range links with reducing the computational cost typically associated with methods like KNN. LSGC overcomes the limitations seen in SVGA by scaling connections logarithmically, improving information flow. We have also proposed LogViG, a hybrid CNN-GNN architecture that integrates LSGC, uses multi-scale and high-resolution features, and leverages a deeper network architecture for enhanced performance. LogViG outperforms existing ViG, CNN, and ViT models on key vision tasks such as image classification and semantic segmentation. Our results demonstrate that LogViG is more efficient in terms of GMACs and parameters while achieving superior accuracy, proving that LSGC and our hybrid architecture offer a significant advancement in the design of Vision GNNs.

\section{Acknowledgements}
\label{Sec:Acknowledgement}

This work is supported in part by the NSF grant CNS 2007284, and in part by the iMAGiNE Consortium (\url{https://imagine.utexas.edu/}).

\newpage

\bibliographystyle{unsrtnat}
\bibliography{reference}

\begin{thebibliography}{53}
\providecommand{\natexlab}[1]{#1}
\providecommand{\url}[1]{\texttt{#1}}
\expandafter\ifx\csname urlstyle\endcsname\relax
  \providecommand{\doi}[1]{doi: #1}\else
  \providecommand{\doi}{doi: \begingroup \urlstyle{rm}\Url}\fi

\bibitem[Krizhevsky et~al.(2012)Krizhevsky, Sutskever, and Hinton]{Alexnet2012}
Alex Krizhevsky, Ilya Sutskever, and Geoffrey~E Hinton.
\newblock Imagenet classification with deep convolutional neural networks.
\newblock \emph{Advances in neural information processing systems}, 25, 2012.

\bibitem[Howard et~al.(2017)Howard, Zhu, Chen, Kalenichenko, Wang, Weyand, Andreetto, and Adam]{howard2017mobilenetsefficientconvolutionalneural}
Andrew~G Howard, Menglong Zhu, Bo~Chen, Dmitry Kalenichenko, Weijun Wang, Tobias Weyand, Marco Andreetto, and Hartwig Adam.
\newblock Mobilenets: Efficient convolutional neural networks for mobile vision applications.
\newblock \emph{arXiv preprint arXiv:1704.04861}, 2017.

\bibitem[He et~al.(2016)He, Zhang, Ren, and Sun]{Resnet}
Kaiming He, Xiangyu Zhang, Shaoqing Ren, and Jian Sun.
\newblock Deep residual learning for image recognition.
\newblock In \emph{Proceedings of the IEEE conference on computer vision and pattern recognition}, pages 770--778, 2016.

\bibitem[Xu et~al.(2021)Xu, Pan, Pan, Hoi, Yi, and Xu]{xu2021regnetselfregulatednetworkimage}
Jing Xu, Yu~Pan, Xinglin Pan, Steven Hoi, Zhang Yi, and Zenglin Xu.
\newblock {RegNet: Self-Regulated Network for Image Classification}, 2021.
\newblock URL \url{https://arxiv.org/abs/2101.00590}.

\bibitem[Tan and Le(2021)]{tan2021efficientnetv2smallermodelsfaster}
Mingxing Tan and Quoc Le.
\newblock Efficientnetv2: Smaller models and faster training.
\newblock In \emph{International conference on machine learning}, pages 10096--10106. PMLR, 2021.

\bibitem[Dosovitskiy et~al.(2021)Dosovitskiy, Beyer, Kolesnikov, Weissenborn, Zhai, Unterthiner, Dehghani, Minderer, Heigold, Gelly, Uszkoreit, and Houlsby]{dosovitskiy2021imageworth16x16words}
Alexey Dosovitskiy, Lucas Beyer, Alexander Kolesnikov, Dirk Weissenborn, Xiaohua Zhai, Thomas Unterthiner, Mostafa Dehghani, Matthias Minderer, Georg Heigold, Sylvain Gelly, Jakob Uszkoreit, and Neil Houlsby.
\newblock {An Image is Worth 16x16 Words: Transformers for Image Recognition at Scale}, 2021.
\newblock URL \url{https://arxiv.org/abs/2010.11929}.

\bibitem[Touvron et~al.(2021)Touvron, Cord, Douze, Massa, Sablayrolles, and J{\'e}gou]{Deit}
Hugo Touvron, Matthieu Cord, Matthijs Douze, Francisco Massa, Alexandre Sablayrolles, and Herv{\'e} J{\'e}gou.
\newblock Training data-efficient image transformers \& distillation through attention.
\newblock In \emph{International conference on machine learning}, pages 10347--10357. PMLR, 2021.

\bibitem[Bao et~al.(2022)Bao, Dong, Piao, and Wei]{bao2022beitbertpretrainingimage}
Hangbo Bao, Li~Dong, Songhao Piao, and Furu Wei.
\newblock {BEiT: BERT Pre-Training of Image Transformers}, 2022.
\newblock URL \url{https://arxiv.org/abs/2106.08254}.

\bibitem[Ronneberger et~al.(2015)Ronneberger, Fischer, and Brox]{ronneberger2015unetconvolutionalnetworksbiomedical}
Olaf Ronneberger, Philipp Fischer, and Thomas Brox.
\newblock {U-Net: Convolutional Networks for Biomedical Image Segmentation}, 2015.
\newblock URL \url{https://arxiv.org/abs/1505.04597}.

\bibitem[Zhang et~al.(2022)Zhang, Tian, Tang, Chu, Wei, Shen, and Liu]{zhang2022segvitsemanticsegmentationplain}
Bowen Zhang, Zhi Tian, Quan Tang, Xiangxiang Chu, Xiaolin Wei, Chunhua Shen, and Yifan Liu.
\newblock {SegViT: Semantic Segmentation with Plain Vision Transformers}, 2022.
\newblock URL \url{https://arxiv.org/abs/2210.05844}.

\bibitem[Han et~al.(2022)Han, Wang, Guo, Tang, and Wu]{Vision_GNN}
Kai Han, Yunhe Wang, Jianyuan Guo, Yehui Tang, and Enhua Wu.
\newblock Vision gnn: An image is worth graph of nodes.
\newblock \emph{arXiv preprint arXiv:2206.00272}, 2022.

\bibitem[Munir et~al.(2023)Munir, Avery, and Marculescu]{MobileViG}
Mustafa Munir, William Avery, and Radu Marculescu.
\newblock Mobilevig: Graph-based sparse attention for mobile vision applications.
\newblock In \emph{Proceedings of the IEEE/CVF Conference on Computer Vision and Pattern Recognition (CVPR) Workshops}, pages 2211--2219, June 2023.

\bibitem[Han et~al.(2023)Han, Wang, Kundu, Ding, and Wang]{ViHGNN}
Yan Han, Peihao Wang, Souvik Kundu, Ying Ding, and Zhangyang Wang.
\newblock Vision hgnn: An image is more than a graph of nodes.
\newblock In \emph{Proceedings of the IEEE/CVF International Conference on Computer Vision}, pages 19878--19888, 2023.

\bibitem[Avery et~al.(2024)Avery, Munir, and Marculescu]{MobileViGv2}
William Avery, Mustafa Munir, and Radu Marculescu.
\newblock Scaling graph convolutions for mobile vision.
\newblock In \emph{Proceedings of the IEEE/CVF Conference on Computer Vision and Pattern Recognition (CVPR) Workshops}, pages 5857--5865, 2024.

\bibitem[Gu et~al.(2022)Gu, Kwon, Wang, Ye, Li, Chen, Lai, Chandra, and Pan]{HRViT}
Jiaqi Gu, Hyoukjun Kwon, Dilin Wang, Wei Ye, Meng Li, Yu-Hsin Chen, Liangzhen Lai, Vikas Chandra, and David~Z Pan.
\newblock Multi-scale high-resolution vision transformer for semantic segmentation.
\newblock In \emph{Proceedings of the IEEE/CVF conference on computer vision and pattern recognition}, pages 12094--12103, 2022.

\bibitem[Deng et~al.(2009)Deng, Dong, Socher, Li, Li, and Fei-Fei]{imagenet1k}
Jia Deng, Wei Dong, Richard Socher, Li-Jia Li, Kai Li, and Li~Fei-Fei.
\newblock Imagenet: A large-scale hierarchical image database.
\newblock In \emph{2009 IEEE Conference on Computer Vision and Pattern Recognition}, pages 248--255, 2009.
\newblock \doi{10.1109/CVPR.2009.5206848}.

\bibitem[Zhou et~al.(2017)Zhou, Zhao, Puig, Fidler, Barriuso, and Torralba]{ADE20K}
Bolei Zhou, Hang Zhao, Xavier Puig, Sanja Fidler, Adela Barriuso, and Antonio Torralba.
\newblock Scene parsing through ade20k dataset.
\newblock In \emph{Proceedings of the IEEE conference on computer vision and pattern recognition}, pages 633--641, 2017.

\bibitem[Jeeveswaran et~al.(2022)Jeeveswaran, Kathiresan, Varma, Magdy, Zonooz, and Arani]{jeeveswaran2022comprehensivestudyvisiontransformers}
Kishaan Jeeveswaran, Senthilkumar Kathiresan, Arnav Varma, Omar Magdy, Bahram Zonooz, and Elahe Arani.
\newblock {A Comprehensive Study of Vision Transformers on Dense Prediction Tasks}, 2022.
\newblock URL \url{https://arxiv.org/abs/2201.08683}.

\bibitem[Thisanke et~al.(2023)Thisanke, Deshan, Chamith, Seneviratne, Vidanaarachchi, and Herath]{thisanke2023semanticsegmentationusingvision}
Hans Thisanke, Chamli Deshan, Kavindu Chamith, Sachith Seneviratne, Rajith Vidanaarachchi, and Damayanthi Herath.
\newblock {Semantic Segmentation using Vision Transformers: A survey}, 2023.
\newblock URL \url{https://arxiv.org/abs/2305.03273}.

\bibitem[Wang et~al.(2022)Wang, Zhang, Wang, and Yang]{10.1145/3508396.3512869}
Xudong Wang, Li~Lyna Zhang, Yang Wang, and Mao Yang.
\newblock {Towards Efficient Vision Transformer Inference: A First Study of Transformers on Mobile Devices}.
\newblock In \emph{Proceedings of the 23rd Annual International Workshop on Mobile Computing Systems and Applications}, HotMobile '22, page 1–7. Association for Computing Machinery, 2022.
\newblock ISBN 9781450392181.

\bibitem[Liu et~al.(2023)Liu, Peng, Zheng, Yang, Hu, and Yuan]{liu2023efficientvitmemoryefficientvision}
Xinyu Liu, Houwen Peng, Ningxin Zheng, Yuqing Yang, Han Hu, and Yixuan Yuan.
\newblock {EfficientViT: Memory Efficient Vision Transformer with Cascaded Group Attention}, 2023.
\newblock URL \url{https://arxiv.org/abs/2305.07027}.

\bibitem[Ma and Zhang(2018)]{ma2018affinitynetsemisupervisedfewshotlearning}
Tianle Ma and Aidong Zhang.
\newblock {AffinityNet: Semi-supervisedFew-shot Learning for Disease Type Prediction}, 2018.
\newblock URL \url{https://arxiv.org/abs/1805.08905}.

\bibitem[Marcheggiani and Titov(2017)]{marcheggiani2017encodingsentencesgraphconvolutional}
Diego Marcheggiani and Ivan Titov.
\newblock Encoding sentences with graph convolutional networks for semantic role labeling.
\newblock \emph{arXiv preprint arXiv:1703.04826}, 2017.

\bibitem[Nguyen and Grishman(2018)]{Nguyen_Grishman_2018}
Thien Nguyen and Ralph Grishman.
\newblock {Graph Convolutional Networks With Argument-Aware Pooling for Event Detection}.
\newblock \emph{Proceedings of the AAAI Conference on Artificial Intelligence}, 32\penalty0 (1), Apr. 2018.
\newblock \doi{10.1609/aaai.v32i1.12039}.
\newblock URL \url{https://ojs.aaai.org/index.php/AAAI/article/view/12039}.

\bibitem[Qiu et~al.(2018)Qiu, Tang, Ma, Dong, Wang, and Tang]{Qiu_2018}
Jiezhong Qiu, Jian Tang, Hao Ma, Yuxiao Dong, Kuansan Wang, and Jie Tang.
\newblock {DeepInf: Social Influence Prediction with Deep Learning}.
\newblock In \emph{Proceedings of the 24th ACM SIGKDD International Conference on Knowledge Discovery \& Data Mining}, KDD ’18. ACM, July 2018.
\newblock \doi{10.1145/3219819.3220077}.
\newblock URL \url{http://dx.doi.org/10.1145/3219819.3220077}.

\bibitem[Monfardini et~al.(2006)Monfardini, Di~Massa, Scarselli, Gori, et~al.]{monfardini2006graph}
Gabriele Monfardini, Vincenzo Di~Massa, Franco Scarselli, Marco Gori, et~al.
\newblock {Graph Neural Networks for Object Localization}.
\newblock \emph{Frontiers in Artificial Intelligence and Applications}, 141:\penalty0 665--669, 2006.

\bibitem[Qi et~al.(2017)Qi, Su, Mo, and Guibas]{qi2017pointnetdeeplearningpoint}
Charles~R Qi, Hao Su, Kaichun Mo, and Leonidas~J Guibas.
\newblock Pointnet: Deep learning on point sets for 3d classification and segmentation.
\newblock In \emph{Proceedings of the IEEE conference on computer vision and pattern recognition}, pages 652--660, 2017.

\bibitem[Bera et~al.(2022)Bera, Wharton, Liu, Bessis, and Behera]{9892682}
Asish Bera, Zachary Wharton, Yonghuai Liu, Nik Bessis, and Ardhendu Behera.
\newblock {SR-GNN: Spatial Relation-Aware Graph Neural Network for Fine-Grained Image Categorization}.
\newblock \emph{IEEE Transactions on Image Processing}, 31:\penalty0 6017--6031, 2022.
\newblock \doi{10.1109/TIP.2022.3205215}.

\bibitem[Jiang et~al.(2023)Jiang, Chen, Tian, and Liu]{10230496}
Juntao Jiang, Xiyu Chen, Guanzhong Tian, and Yong Liu.
\newblock {ViG-UNet: Vision Graph Neural Networks for Medical Image Segmentation}.
\newblock In \emph{{2023 IEEE 20th International Symposium on Biomedical Imaging (ISBI)}}, pages 1--5, 2023.
\newblock \doi{10.1109/ISBI53787.2023.10230496}.

\bibitem[Shou et~al.(2024)Shou, Ai, Meng, and Yin]{shou2024graphinformationbottleneckremote}
Yuntao Shou, Wei Ai, Tao Meng, and Nan Yin.
\newblock {Graph Information Bottleneck for Remote Sensing Segmentation}, 2024.
\newblock URL \url{https://arxiv.org/abs/2312.02545}.

\bibitem[Alon and Yahav(2020)]{alon2020bottleneck}
Uri Alon and Eran Yahav.
\newblock On the bottleneck of graph neural networks and its practical implications.
\newblock \emph{arXiv preprint arXiv:2006.05205}, 2020.

\bibitem[Wang et~al.(2020)Wang, Sun, Cheng, Jiang, Deng, Zhao, Liu, Mu, Tan, Wang, et~al.]{HRNet}
Jingdong Wang, Ke~Sun, Tianheng Cheng, Borui Jiang, Chaorui Deng, Yang Zhao, Dong Liu, Yadong Mu, Mingkui Tan, Xinggang Wang, et~al.
\newblock Deep high-resolution representation learning for visual recognition.
\newblock \emph{IEEE transactions on pattern analysis and machine intelligence}, 43\penalty0 (10):\penalty0 3349--3364, 2020.

\bibitem[Yu et~al.(2021)Yu, Xiao, Gao, Yuan, Zhang, Sang, and Wang]{Lite_HRNet}
Changqian Yu, Bin Xiao, Changxin Gao, Lu~Yuan, Lei Zhang, Nong Sang, and Jingdong Wang.
\newblock Lite-hrnet: A lightweight high-resolution network.
\newblock In \emph{Proceedings of the IEEE/CVF conference on computer vision and pattern recognition}, pages 10440--10450, 2021.

\bibitem[Nguyen et~al.(2020)Nguyen, Raghu, and Kornblith]{nguyen2020wide}
Thao Nguyen, Maithra Raghu, and Simon Kornblith.
\newblock Do wide and deep networks learn the same things? uncovering how neural network representations vary with width and depth.
\newblock \emph{arXiv preprint arXiv:2010.15327}, 2020.

\bibitem[Liu et~al.(2022)Liu, Mao, Wu, Feichtenhofer, Darrell, and Xie]{liu2022convnet}
Zhuang Liu, Hanzi Mao, Chao-Yuan Wu, Christoph Feichtenhofer, Trevor Darrell, and Saining Xie.
\newblock A convnet for the 2020s.
\newblock In \emph{Proceedings of the IEEE/CVF Conference on Computer Vision and Pattern Recognition}, pages 11976--11986, 2022.

\bibitem[Li et~al.(2022)Li, Yuan, Wen, Hu, Evangelidis, Tulyakov, Wang, and Ren]{EfficientFormer}
Yanyu Li, Geng Yuan, Yang Wen, Ju~Hu, Georgios Evangelidis, Sergey Tulyakov, Yanzhi Wang, and Jian Ren.
\newblock Efficientformer: Vision transformers at mobilenet speed.
\newblock \emph{Advances in Neural Information Processing Systems}, 35:\penalty0 12934--12949, 2022.

\bibitem[Graham et~al.(2021)Graham, El-Nouby, Touvron, Stock, Joulin, J{\'e}gou, and Douze]{graham2021levit}
Benjamin Graham, Alaaeldin El-Nouby, Hugo Touvron, Pierre Stock, Armand Joulin, Herv{\'e} J{\'e}gou, and Matthijs Douze.
\newblock Levit: a vision transformer in convnet's clothing for faster inference.
\newblock In \emph{Proceedings of the IEEE/CVF international conference on computer vision}, pages 12259--12269, 2021.

\bibitem[Wang et~al.(2021)Wang, Xie, Li, Fan, Song, Liang, Lu, Luo, and Shao]{wang2021pyramid}
Wenhai Wang, Enze Xie, Xiang Li, Deng-Ping Fan, Kaitao Song, Ding Liang, Tong Lu, Ping Luo, and Ling Shao.
\newblock Pyramid vision transformer: A versatile backbone for dense prediction without convolutions.
\newblock In \emph{Proceedings of the IEEE/CVF international conference on computer vision}, pages 568--578, 2021.

\bibitem[Liu et~al.(2021)Liu, Lin, Cao, Hu, Wei, Zhang, Lin, and Guo]{liu2021swin}
Ze~Liu, Yutong Lin, Yue Cao, Han Hu, Yixuan Wei, Zheng Zhang, Stephen Lin, and Baining Guo.
\newblock Swin transformer: Hierarchical vision transformer using shifted windows.
\newblock In \emph{Proceedings of the IEEE/CVF international conference on computer vision}, pages 10012--10022, 2021.

\bibitem[Yu et~al.(2022)Yu, Luo, Zhou, Si, Zhou, Wang, Feng, and Yan]{MetaFormer}
Weihao Yu, Mi~Luo, Pan Zhou, Chenyang Si, Yichen Zhou, Xinchao Wang, Jiashi Feng, and Shuicheng Yan.
\newblock Metaformer is actually what you need for vision.
\newblock In \emph{Proceedings of the IEEE/CVF Conference on Computer Vision and Pattern Recognition}, pages 10819--10829, 2022.

\bibitem[Paszke et~al.(2019)]{paszke2019pytorch}
Adam Paszke et~al.
\newblock Pytorch: An imperative style, high-performance deep learning library.
\newblock \emph{Advances in neural information processing systems}, 32, 2019.

\bibitem[Wightman(2019)]{timm}
Ross Wightman.
\newblock {PyTorch Image Models}.
\newblock \url{https://github.com/rwightman/pytorch-image-models}, 2019.

\bibitem[Loshchilov and Hutter(2017)]{AdamW}
Ilya Loshchilov and Frank Hutter.
\newblock Decoupled weight decay regularization.
\newblock \emph{arXiv preprint arXiv:1711.05101}, 2017.

\bibitem[Munir et~al.(2024)Munir, Avery, Rahman, and Marculescu]{GreedyViG}
Mustafa Munir, William Avery, Md~Mostafijur Rahman, and Radu Marculescu.
\newblock Greedyvig: Dynamic axial graph construction for efficient vision gnns.
\newblock In \emph{Proceedings of the IEEE/CVF Conference on Computer Vision and Pattern Recognition}, pages 6118--6127, 2024.

\bibitem[Li et~al.(2023)Li, Hu, Wen, Evangelidis, Salahi, Wang, Tulyakov, and Ren]{EfficientFormerv2}
Yanyu Li, Ju~Hu, Yang Wen, Georgios Evangelidis, Kamyar Salahi, Yanzhi Wang, Sergey Tulyakov, and Jian Ren.
\newblock Rethinking vision transformers for mobilenet size and speed.
\newblock In \emph{Proceedings of the IEEE/CVF International Conference on Computer Vision}, pages 16889--16900, 2023.

\bibitem[Radosavovic et~al.(2020)Radosavovic, Kosaraju, Girshick, He, and Doll{\'a}r]{RegNetY}
Ilija Radosavovic, Raj~Prateek Kosaraju, Ross Girshick, Kaiming He, and Piotr Doll{\'a}r.
\newblock Designing network design spaces.
\newblock In \emph{Proceedings of the IEEE/CVF conference on computer vision and pattern recognition}, pages 10428--10436, 2020.

\bibitem[Cubuk et~al.(2020)Cubuk, Zoph, Shlens, and Le]{RandAugment}
Ekin~D Cubuk, Barret Zoph, Jonathon Shlens, and Quoc~V Le.
\newblock Randaugment: Practical automated data augmentation with a reduced search space.
\newblock In \emph{Proceedings of the IEEE/CVF conference on computer vision and pattern recognition workshops}, pages 702--703, 2020.

\bibitem[Zhang et~al.(2018)Zhang, Cisse, Dauphin, and Lopez-Paz]{Mixup}
Hongyi Zhang, Moustapha Cisse, Yann~N. Dauphin, and David Lopez-Paz.
\newblock mixup: Beyond empirical risk minimization.
\newblock In \emph{International Conference on Learning Representations}, 2018.
\newblock URL \url{https://openreview.net/forum?id=r1Ddp1-Rb}.

\bibitem[Yun et~al.(2019)Yun, Han, Oh, Chun, Choe, and Yoo]{CutMix}
Sangdoo Yun, Dongyoon Han, Seong~Joon Oh, Sanghyuk Chun, Junsuk Choe, and Youngjoon Yoo.
\newblock Cutmix: Regularization strategy to train strong classifiers with localizable features.
\newblock In \emph{Proceedings of the IEEE/CVF international conference on computer vision}, pages 6023--6032, 2019.

\bibitem[Zhong et~al.(2020)Zhong, Zheng, Kang, Li, and Yang]{RandomErase}
Zhun Zhong, Liang Zheng, Guoliang Kang, Shaozi Li, and Yi~Yang.
\newblock Random erasing data augmentation.
\newblock In \emph{Proceedings of the AAAI conference on artificial intelligence}, volume~34, pages 13001--13008, 2020.

\bibitem[Hoffer et~al.(2020)Hoffer, Ben-Nun, Hubara, Giladi, Hoefler, and Soudry]{RepeatedAugment}
Elad Hoffer, Tal Ben-Nun, Itay Hubara, Niv Giladi, Torsten Hoefler, and Daniel Soudry.
\newblock Augment your batch: Improving generalization through instance repetition.
\newblock In \emph{Proceedings of the IEEE/CVF Conference on Computer Vision and Pattern Recognition}, pages 8129--8138, 2020.

\bibitem[Kirillov et~al.(2019)Kirillov, Girshick, He, and Doll{\'a}r]{kirillov2019panoptic}
Alexander Kirillov, Ross Girshick, Kaiming He, and Piotr Doll{\'a}r.
\newblock Panoptic feature pyramid networks.
\newblock In \emph{Proceedings of the IEEE/CVF conference on computer vision and pattern recognition}, pages 6399--6408, 2019.

\bibitem[Vasu et~al.(2023)Vasu, Gabriel, Zhu, Tuzel, and Ranjan]{FastViT}
Pavan Kumar~Anasosalu Vasu, James Gabriel, Jeff Zhu, Oncel Tuzel, and Anurag Ranjan.
\newblock Fastvit: A fast hybrid vision transformer using structural reparameterization.
\newblock In \emph{Proceedings of the IEEE/CVF International Conference on Computer Vision (ICCV)}, pages 5785--5795, October 2023.

\end{thebibliography}

\end{document}